\documentclass[conference]{IEEEtran}
\IEEEoverridecommandlockouts

\usepackage{cite}
\usepackage{amsmath,amssymb,amsfonts}
\usepackage{algorithm}
\usepackage{algorithmic}
\usepackage{placeins}
\usepackage{graphicx}
\usepackage{textcomp}
\usepackage{xcolor}
\usepackage{titlesec}
\usepackage[utf8x]{inputenc}
\usepackage{subcaption}
\usepackage{stfloats}
\usepackage{textcomp}
\usepackage{float}  
\usepackage{hyperref}
\hypersetup{
    colorlinks=true,
    linkcolor=red,
    filecolor=magenta,      
    urlcolor=blue,
    pdftitle={Overleaf Example},
    pdfpagemode=FullScreen,
    }
\begin{document}

\title{Boosting Imperceptibility of Stable Diffusion-based Adversarial Examples Generation with Momentum}

\author{
Nashrah Haque\IEEEauthorrefmark{1}, Xiang Li\IEEEauthorrefmark{1},  Zhehui Chen\IEEEauthorrefmark{2},
Yanzhao Wu\IEEEauthorrefmark{3}, Lei Yu\IEEEauthorrefmark{4}, 
Arun Iyengar\IEEEauthorrefmark{5},
Wenqi Wei\IEEEauthorrefmark{1} 
\\

\IEEEauthorblockA{\IEEEauthorrefmark{1} Fordham University, New York, NY, USA, \{nhaque14, xl5, wwei23\}@fordham.edu}
\IEEEauthorblockA{\IEEEauthorrefmark{2} 
Google, Mountain View, California, USA, zchen451@gatech.edu}
\IEEEauthorblockA{\IEEEauthorrefmark{3} Florida International University, Miami, FL, USA, yawu@fiu.edu }
\IEEEauthorblockA{\IEEEauthorrefmark{4} Rensselaer Polytechnic Institute, Troy, NY, USA, yul9@rpi.edu }
\IEEEauthorblockA{\IEEEauthorrefmark{5} Cisco Research, San Jose, CA, USA, aruniyen3@gmail.com }
}

\maketitle
\thispagestyle{plain}

\begin{abstract}
We propose a novel framework, Stable Diffusion-based Momentum Integrated Adversarial Examples (SD-MIAE), for generating adversarial examples that can effectively mislead neural network classifiers while maintaining visual imperceptibility and preserving the semantic similarity to the original class label. Our method leverages the text-to-image generation capabilities of the Stable Diffusion model by manipulating token embeddings corresponding to the specified class in its latent space. These token embeddings guide the generation of adversarial images that maintain high visual fidelity. The SD-MIAE framework consists of two phases: (1) an initial adversarial optimization phase that modifies token embeddings to produce misclassified yet natural-looking images and (2) a momentum-based optimization phase that refines the adversarial perturbations. By introducing momentum, our approach stabilizes the optimization of perturbations across iterations, enhancing both the misclassification rate and visual fidelity of the generated adversarial examples. Experimental results demonstrate that SD-MIAE achieves a high misclassification rate of 79\%, improving by 35\% over the state-of-the-art method while preserving the imperceptibility of adversarial perturbations and the semantic similarity to the original class label, making it a practical method for robust adversarial evaluation.
\end{abstract}


\section{Introduction}

Deep neural networks (DNNs) have achieved remarkable success across various domains, including image classification~\cite{b1}, speech recognition~\cite{b2}, and natural language processing~\cite{b3}. These advances are primarily attributable to the ability of DNNs to learn complex patterns from vast datasets, enabling them to outperform traditional methods in numerous tasks. However, despite these achievements, DNNs are inherently vulnerable to adversarial attacks~\cite{b4}—small, often imperceptible perturbations to input data that can lead to significant misclassifications and pose serious security risks. These vulnerabilities are particularly concerning in safety-critical applications such as autonomous driving~\cite{b5} and healthcare~\cite{b6}, where the consequences of model failures can be catastrophic. The ability of adversarial examples to exploit these model weaknesses underscores the urgent need for methods to generate human yet deceptive inputs capable of fooling models while evading detection by automated systems and human observers.

Recent advancements in generative models, particularly text-to-image diffusion models like Stable Diffusion~\cite{b19}, have introduced a new dimension in the field of adversarial attacks. These models, known for their ability to generate highly realistic images from textual descriptions, have rapidly gained popularity across various creative and industrial applications. However, as their sophistication and usage grow, they have become prime targets for adversarial attacks~\cite{b27}. These attacks aim to subtly manipulate the generated outputs, leading to misclassification or other unintended outcomes while maintaining high visual fidelity~\cite{b28}. Here, visual fidelity refers to the adversarial perturbation being imperceptible, with the generated image maintaining semantic alignment with the class label described in the prompts. 

However, achieving this balance remains a major challenge, especially for adversarial example generation methods based on models like Stable Diffusion~\cite{b12}.
The process of perturbing token embeddings in high-dimensional latent spaces during image generation often results in images with unnatural artifacts, which compromise the adversarial attack's effectiveness by making the perturbations more detectable~\cite{b12}. This highlights a broader challenge in the field: balancing the subtlety of adversarial perturbations with the need to preserve the natural appearance and semantic similarity to the original text prompts of the generated content. This challenge has been echoed in recent literature, particularly in the context of text-to-image diffusion models~\cite{b27}.

\begin{figure*}[t]
    \centering
    \begin{subfigure}{0.48\textwidth}
        \centering
        \includegraphics[width=\textwidth]{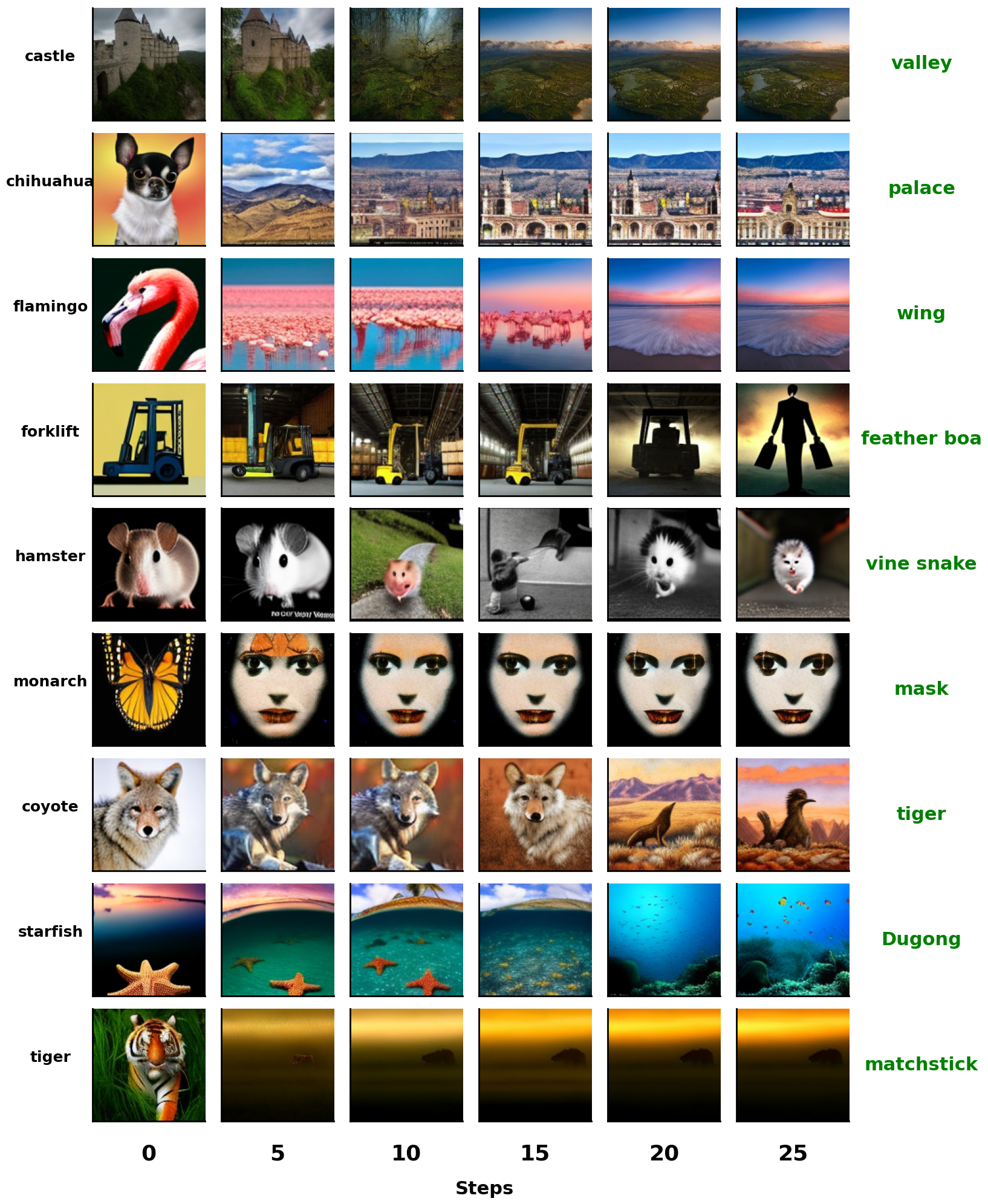}
        \caption{SD-NAE}
        \label{fig:sub1}
    \end{subfigure}
    \hspace{0.02\textwidth} 
    \begin{subfigure}{0.48\textwidth}
        \centering
        \includegraphics[width=\textwidth]{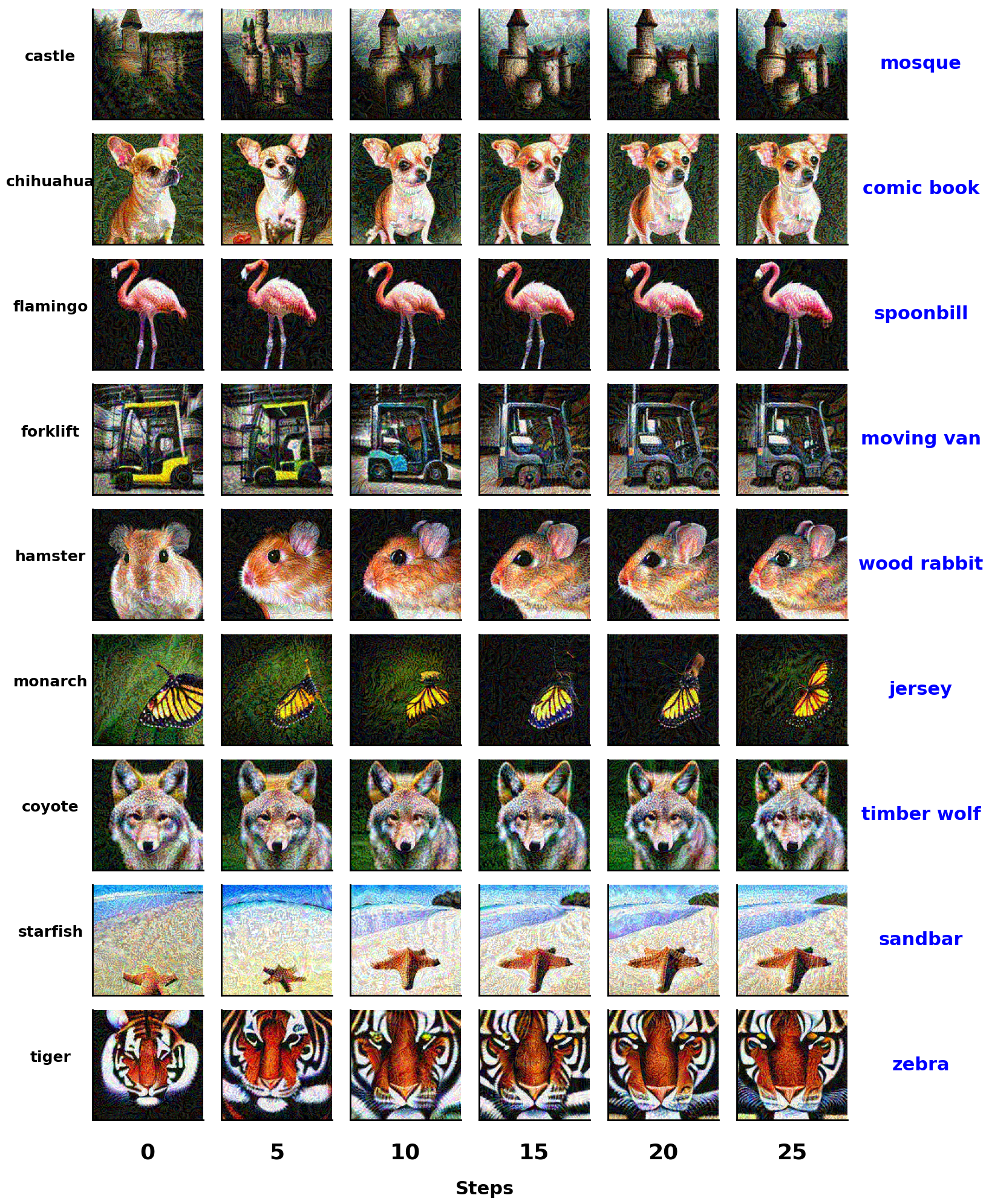}
        \caption{SD-MIAE}
        \label{fig:sub2}
    \end{subfigure}
    \caption{Qualitative comparison of adversarial examples generated by SD-NAE and SD-MIAE. As iteration steps increase, the semantic contents of adversarial examples generated by SD-NAE gradually deviate from their original class label, becoming easily detectable by human observers. In contrast, SD-MIAE produces adversarial examples that remain visually imperceptible and semantically aligned with the original class label.}
    \label{fig:comparison}
\vspace{-0.4cm}
\end{figure*}

To address these challenges, and motivated by recent findings~\cite{b27,b28}, which emphasize the importance of stabilizing perturbations in adversarial example generation, particularly when working with complex generative models like diffusion models, we propose \textbf{Stable Diffusion-based Momentum Integrated Adversarial Examples (SD-MIAE)}. This novel framework builds upon~\cite{b12}
by incorporating momentum-based optimization techniques. 

The momentum-based optimization in the SD-MIAE framework is introduced after the initial image generation process. Specifically, we first iteratively perturb token embeddings to generate adversarial examples and then employ a momentum-based refinement technique to maintain control over the adversarial modifications, ensuring their effectiveness. This process enables SD-MIAE to produce adversarial examples that are not only more effective at inducing misclassification but also visually imperceptible and semantically similar to the original class label, thereby reducing detectable artifacts. 

\textbf{Figure~\ref{fig:comparison}} illustrates the adversarial examples generated by SD-MIAE and SD-NAE~\cite{b12}. As iteration steps increase, the semantic contents of adversarial examples generated by SD-NAE gradually deviate from their original class label, becoming easily detectable by human observers. In contrast, SD-MIAE produces adversarial examples that remain visually imperceptible and semantically consistent with the original class label, highlighting its ability to generate adversarial examples with high visual fidelity while causing misclassifications.


In addition to enhancing the quality and effectiveness of adversarial examples, SD-MIAE introduces opportunities for more robust evaluation of real-world systems~\cite{b16}. SD-MIAE addresses this challenge by generating examples that mislead classifiers and remain imperceptible to human observers. This characteristic is crucial when adversarial attacks could compromise security systems or decision-making processes without raising suspicion~\cite{b33}. This approach enables SD-MIAE to provide a versatile framework for testing the resilience of AI-driven systems against subtle yet highly effective perturbations.

Our contributions are summarized as follows:
\begin{itemize}
    \item \textbf{Momentum-based optimization in Stable Diffusion:} We introduce momentum into the optimization of adversarial perturbations, significantly enhancing the stability of perturbations across iterations. This reduces artificial artifacts and maintains the natural appearance of adversarial examples and their semantic similarity to the original class labels, addressing a key limitation identified in previous work~\cite{b27,b12}.
    \item \textbf{Improved attack performance:} Extensive experiments are conducted to verify the effectiveness of SD-MIAE, which achieved a misclassification rate of 79\%, improving by 35\% over the state-of-the-art method~\cite{b12}. This demonstrates the effectiveness of momentum-based optimization in generating effective adversarial examples that are challenging for classifiers to correctly classify.
    \item \textbf{Fine-tuned control over perturbations}: Our method allows for precise adjustment of the perturbation magnitude \( \epsilon \) and momentum factor \( \mu \), providing flexibility in balancing the strength of the adversarial perturbations and maintaining image quality. This ensures that even small, controlled perturbations can induce misclassification while preserving the natural appearance of the images.

\end{itemize}

\section{Related Work}

This section reviews recent advancements in adversarial example generation, focusing on adversarial attacks in text-to-image models, Natural Adversarial Examples (NAEs), and the application of Stable Diffusion models. We also discuss the limitations of current approaches and how they relate to our work.

\subsection{Adversarial Attacks}
\label{subsec:adversarial_attack}
Adversarial robustness can be significantly improved through techniques like adversarial training, where models are fine-tuned using adversarial examples to resist attacks \cite{b14}.Research has shown that adversarial examples exploit the inherent linear nature of neural networks\cite{b8}, leading to vulnerabilities that can be efficiently exploited using methods such as gradient-based attacks~\cite{b4,b14}. These findings laid the groundwork for using adversarial examples as a tool for enhancing model robustness through adversarial training. Additionally, methods for generating adversarial examples that are both highly effective and difficult to detect have further emphasized the challenges in defending against such sophisticated attacks~\cite{b16}. In parallel, other attack strategies introduce visible or subtle perturbations, such as adversarial patches~\cite{b20}, while defense mechanisms like feature squeezing~\cite{b21} aim to simplify input features to detect and mitigate these attacks.

Adversarial attacks in the context of text-to-image models involve perturbing the input prompt in various ways to mislead the model into generating incorrect or malicious outputs. Based on the granularity of perturbations, existing attacks can be primarily categorized into three levels: character-level, word-level, and sentence-level, depending on how the adversarial examples are generated~\cite{b29}.

\textbf{Character-level perturbations} involve altering, adding, or removing characters within a word. These subtle changes can be difficult to detect yet effective in misleading models.

\textbf{Word-level perturbations} aim to manipulate entire words within the prompt. This includes replacing, inserting, or deleting words, which can significantly alter the generated image's content.

\textbf{Sentence-level perturbations} involve rewriting or rephrasing entire sentences within the prompt, which can introduce more complex and comprehensive changes to the generated images.

Adversarial attacks on text-to-image models can be classified along three key dimensions: the target of the attack (untargeted vs. targeted), the adversary's knowledge of the system (white-box vs. black-box), and the type of perturbation applied (character, word, or sentence level). These categories provide a framework for understanding how adversarial prompts are generated and their implications on the robustness of such models \cite{b27}.

\subsection{Natural Adversarial Examples (NAEs)}
Natural Adversarial Examples (NAEs) are a class of adversarial examples that arise naturally in the data without requiring artificial perturbations. NAEs are defined as a set of real-world samples with respect to a target classifier $F$~\cite{b9,b15}:

\begin{equation}
A \equiv \{x \in S \mid O(x) \neq F(x)\}
\end{equation}

Where $A$ denotes the set of NAEs that are misclassified by the model, $S$ represents all images that naturally occur and are realistic to human observers, $O(x)$ is the true label assigned by the model, and $F(x)$ is the predicted label of image $x$.

Hendrycks et al. \cite{b9} are among the first to explore NAEs systematically, demonstrating that naturally occurring examples could effectively reveal the weaknesses of deep learning models. These examples are particularly valuable in assessing model robustness, as they reflect the types of challenges that models are likely to encounter in real-world scenarios.

Unlike NAEs, which may arise naturally without any deliberate modification, traditional adversarial examples involve pixel-level perturbations carefully crafted to mislead classifiers using multiple gradient-based methods \cite{b14}.

\subsection{Stable Diffusion}
Stable Diffusion~\cite{b19} is a class of latent diffusion models designed for conditional generation tasks such as text-to-image synthesis. These models generate high-quality images based on textual descriptions by transforming a random latent vector $z$ and a corresponding text embedding $e_{\text{text}}$ into an image:
\begin{equation}
x = G(z; e_{\text{text}})
\end{equation}
where $G$ represents the generative process, and $e_{\text{text}}$ is often derived from a transformer-based text encoder\cite{b34}. This encoder processes the input textual description and output token embeddings used to guide the image generation.

Despite their effectiveness, Stable Diffusion models exhibit vulnerabilities, particularly when subjected to adversarial attacks, as mentioned in Section~\ref{subsec:adversarial_attack}. 
These vulnerabilities highlight the need for more robust defenses against adversarial manipulations.

\subsection{Recent Contributions and Limitations}

Recent advancements in generating adversarial examples have leveraged Stable Diffusion models by utilizing custom embeddings to fine-tune the diffusion process, optimizing the generation of adversarial examples. These methods perturb only the condition embedding without altering the underlying sampling process, which helps generalize across various diffusion models \cite{b12}. However, challenges remain in balancing the perturbations to retain the visual fidelity of generated adversarial examples and ensure adaptability across different model architectures. Previous work~\cite{dhariwal2021diffusion, ho2021classifier} enforces adversarial classifier guidance, requiring significant modifications to classifier-free guidance sampling, complicating adaptation to different samplers.

Momentum-based optimization has emerged as a powerful technique to enhance the effectiveness of adversarial attacks by stabilizing the gradient updates across iterations, as demonstrated by Dong et al.~\cite{b7}. \cite{b7} introduces a momentum term into the iterative optimization of generating adversarial examples, demonstrating significant improvements in attack success rates while maintaining a lower computational cost. It not only addresses some of the instability issues observed in earlier methods but also provides a framework that can be adapted to various adversarial settings, including the generation of visually coherent adversarial examples.

Built on these insights, our research seeks to further explore the potential of generative models in producing effective and visually coherent adversarial examples, addressing some of the key challenges identified in previous studies~\cite{b27,b12}.

\section{Methodology}

\begin{figure}[t]
    \centering
    \includegraphics[width=1.0\linewidth]{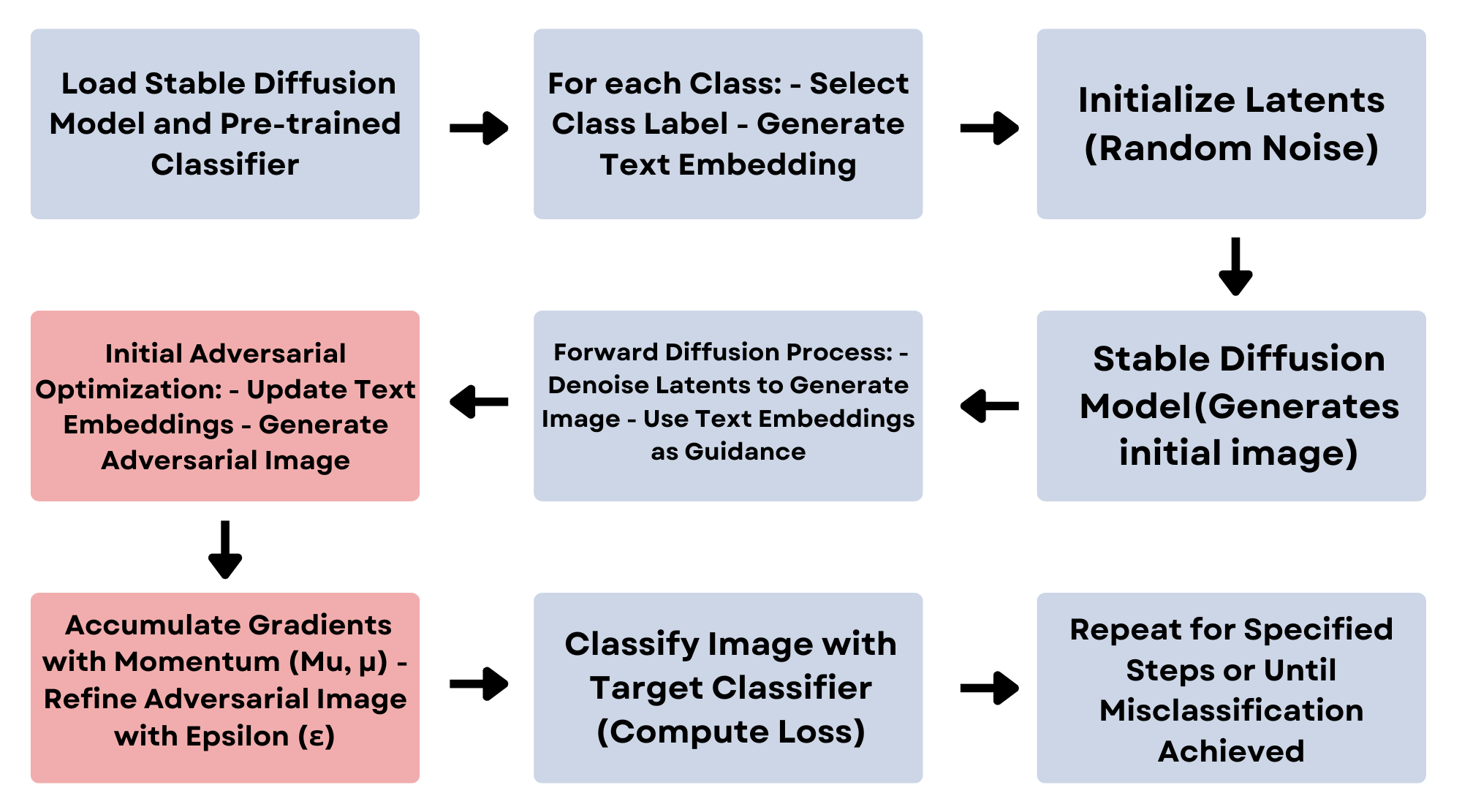}
    \caption{SD-MIAE workflow. Key steps include loading the models, generating text embeddings, performing forward diffusion, applying the SD-MIAE adversarial attack, classifying the image, and optimizing the token embeddings through backpropagation. The process is iteratively repeated to refine the adversarial example until misclassification is achieved.}
    \label{fig:sdnae_mifgsm}
\end{figure}

In this work, we introduce a novel framework, Stable Diffusion-based Momentum Integrated Adversarial Examples (SD-MIAE), which is designed to generate adversarial examples that can effectively mislead neural network classifiers while maintaining visually imperceptible and preserving the semantic similarity to the original class label. SD-MIAE leverages the text-to-image generation capabilities of the Stable Diffusion model, manipulating the latent space through token embeddings to create images that can be misclassified by the target model. The generated adversarial examples are further refined using momentum-based optimization, which enhances the effectiveness of the perturbations. \textbf{Figure~\ref{fig:sdnae_mifgsm}} illustrates the workflow of SD-MIAE.

\subsection{Threat Model}
Stable Diffusion-based Adversarial examples generation fits into the following threat scenario: An attacker exploits an open-sourced image classifier (e.g., ResNet50) and a generative model (e.g. Stable Diffusion) to generate adversarial examples that mislead the classifier to predict them into any other classes (i.e., untargeted attack) while preserving semantic similarities to their original class labels. We assume the attacker has full knowledge of both the image classifier and generative model so they can achieve the attack by manipulating both the image generation process and generated images. Since these adversarial images are visually indistinguishable from clean images and maintain their semantic similarities, they are challenging to detect by human inspectors or existing defense mechanisms and can lead to malicious classifier behavior when deployed in real-world settings.

\subsection{Generating Adversarial Examples}

The SD-MIAE framework begins by generating adversarial examples that retain a natural appearance while being effective at misleading a classifier. This process involves optimizing token embeddings associated with the textual description of the target class. These token embeddings, representing key semantic attributes of the text prompt within the latent space of the Stable Diffusion, serve as the foundation for generating adversarial images.

\subsubsection{Token Embedding Initialization}

The SD-MIAE process starts by converting a textual description—such as ``A high-quality image of a hamster"—into a set of token embeddings, which encodes the semantic content of the text into a high-dimension space. The embeddings represent critical aspects of the text prompt and are essential in guiding the image generation process. By optimizing the token embeddings associated with the class labels, SD-MIAE manipulates the underlying semantics of the generated image, making it adversarial.

\subsubsection{Latent Vector Initialization}

Once the token embeddings are prepared, random latent vectors $ z $ are initialized. These latent vectors are crucial as they provide the necessary diversity to the generation process, allowing the Stable Diffusion model to produce diverse outputs. The model processes the vectors and synthesizes an initial image that closely aligns with the given text prompt. This image typically represents the intended class accurately and is expected to be correctly classified by the target classifier.

\subsection{Initial Adversarial Optimization}

The core of our methodology revolves around an initial adversarial optimization process, where the goal is to modify the token embeddings associated with the class labels to produce an adversarial image. During this phase, the optimization focuses solely on the token embeddings without applying momentum $\mu$ or bound for perturbation $ \epsilon$. The token embeddings are iteratively updated to create an image that remains visually consistent with the original text prompt but induces misclassification by the classifier $F$.

At each iteration $t$, the current set of token embeddings is used to generate an image via the Stable Diffusion. This image is then passed to the classifier, and a loss is computed based on the following components:

\begin{itemize}
    \item \textbf{Adversarial Loss:} This loss is designed to cause the classifier to misclassify the generated examples. For targeted attacks, where the goal is to induce misclassification towards a specified class $y^*$ ($y*\neq y$), cross-entropy can be used as the adversarial loss. For untargeted attack, one can use negative cross-entropy as the adversarial loss (i.e., maximizing the classification loss of the original label $y$).
    \item \textbf{Cosine Similarity Regularization:} To ensure that the generated image remains visually similar to the original prompt, a cosine similarity regularization term is employed. This term penalizes significant deviations in the token embeddings, thus preserving the natural appearance of the image while making subtle, adversarial alterations.
\end{itemize}

The combined objective for untargeted attack is expressed as follow:

\begin{equation}
\label{Equation:loss}
\begin{split}
\min -\ell(F(G(z; e_{\text{text}})), y) + \lambda \cdot R(\hat{e}_{\text{token}}^k, e_{\text{token}}^k), \\
\text{where } e_{\text{text}} = E(e_{\text{token}}^0, \ldots, \hat{e}_{\text{token}}^k, \ldots, e_{\text{token}}^{K-1})
\end{split}
\end{equation}

where $e_{\text{text}}$ denotes the perturbed text embeddings used to guide the image generation, $e_{\text{token}}^k$ denotes the token embeddings associated with the class label, $\lambda$ is the coefficient of regularization that balances the trade-off between causing misclassification and maintaining visual fidelity, and $ R(e_{\text{token}}^*, e_{\text{token}})$ represents the cosine similarity between the original and perturbed embeddings, serving as a regularization term.


The token embeddings are optimized iteratively until the desired adversarial effect is achieved. This phase lays the groundwork for creating an adversarial example, but it operates without the additional stability and refinement introduced by momentum.

\subsection{Momentum-Based Optimization}
\label{method:momentum}

After the initial adversarial optimization, SD-MIAE employs a momentum-based optimization technique to further refine the adversarial example. This approach builds upon the foundation laid during the initial optimization by introducing momentum $ \mu $ and bounding for the magnitude of perturbation $ \epsilon $ to stabilize and enhance the 
effectiveness of the adversarial attack. The momentum-based method ensures that the perturbations applied to the image are consistent across iterations, effectively guiding the optimization process out of local minima and towards a more effective adversarial example.

\subsubsection{Initialization of Momentum}

The momentum-based optimization begins with the adversarial image generated from the initial phase. At this point, the momentum term $ m_0 $ is initialized to zero, indicating that no prior gradient information is being carried over. The process is now ready to refine the adversarial perturbation through iterative updates.

\subsubsection{Gradient Accumulation and Update}

During each iteration, the gradient of the loss function with respect to the generated image is computed. This gradient indicates the direction in which the image should be perturbed to increase the confidence of misclassification by the classifier $ F $. Rather than applying this gradient directly, SD-MIAE accumulates it using a momentum term $ m_t $, which combines the current gradient with the momentum from previous iterations:

\begin{equation}
\label{equa:momentum}
m_{t+1} = \mu \cdot m_t + \frac{\nabla_x \ell(F(x_t), y)}{\|\nabla_x \ell(F(x_t), y)\|_1}
\end{equation}

where $ \nabla_x \ell(F(x_t), y) $ represents the gradient of the loss with respect to the image at iteration $ t $. The momentum term $ m_t $ smooths the perturbations applied to the image, leading to a more stable and effective adversarial attack.

\subsubsection{Perturbation of the Adversarial Image}

Following the update of the momentum term $ m_t $, the adversarial image $x_t $ is adjusted by adding a perturbation in the direction indicated by the accumulated momentum:

\begin{equation}
x_{t+1} = x_t + \alpha \cdot \text{sign}(m_{t+1})
\end{equation}

where $\alpha$ is the step size and set to  $\epsilon/T$ to make the generated adversarial examples satisfy the $L_\infty$ bound. $\epsilon$ and $T$ denote the perturbation size and total iteration steps, respectively. $ m_{t+1} $ is the momentum term that has been accumulated and updated throughout the iterative process. This step ensures that the adversarial image is progressively refined while minimizing unnatural distortions caused by the perturbations.

\subsubsection{Iterative Refinement and Final Output}

The optimization of the adversarial perturbation is repeated over a series of iterations. The use of momentum helps to guide the optimization process toward a highly effective adversarial example, avoiding common pitfalls such as gradient masking or entrapment in local minima, which are typical challenges in traditional gradient-based attacks.

Upon completing the momentum-based optimization process, the final adversarial image is produced. This image integrates the adversarial characteristics from the initial optimization phase with the refined perturbations achieved through momentum-based updates, resulting in adversarial examples that can effectively mislead the target classifier.

\begin{algorithm}[t]
\caption{SD-MIAE: Stable Diffusion-based Momentum-Integrated Adversarial Examples}
\label{algo:SD-MIAE}
\begin{algorithmic}[1]
\REQUIRE Classifier $F$, Stable Diffusion model $G$, Initial token embeddings $e_{\text{token}}$, Perturbation size $\epsilon$, Momentum factor $\mu$, Learning rate $\eta$, Number of embedding optimization steps $T_{\text{embed}}$, Number of attack iterations $T_{\text{attack}}$
\ENSURE Adversarial image $x^*$ such that $\|x^* - x_0\|_\infty \leq \epsilon$

\STATE Set step size $\alpha = \epsilon / T_{\text{attack}}$

\FOR{each embedding optimization step $t = 1$ to $T_{\text{embed}}$}
    \STATE Generate image $x_0 = G(z; e_{\text{token}})$
    \STATE Compute classifier output $F(x_0)$
    \STATE Compute adversarial loss: $\ell_{\text{adv}} = -\ell(F(x_0), y)$
    \STATE Update token embeddings: $e_{\text{token}} \leftarrow e_{\text{token}} - \eta \nabla_{e_{\text{token}}} \ell_{\text{adv}}$
    \STATE \textbf{Momentum-based refinement:}
    \STATE Initialize adversarial image $x = x_0$
    \STATE Initialize momentum $m = 0$
    \FOR{each attack iteration $k = 1$ to $T_{\text{attack}}$}
        \STATE Compute gradient $g = \nabla_x \ell(F(x), y)$
        \STATE Normalize gradient: $g = \dfrac{g}{\|g\|_1 + \delta}$ \hfill (small $\delta$ to avoid division by zero)
        \STATE Update momentum: $m = \mu \cdot m + g$
        \STATE Update image: $x = x + \alpha \cdot \text{sign}(m)$
        \STATE Project $x$ onto $\epsilon$-ball around $x_0$: $\|x - x_0\|_\infty \leq \epsilon$
        \STATE Clip $x$ to valid pixel range $[0, 1]$
    \ENDFOR
\ENDFOR

\STATE Return final adversarial image $x^* = x$

\end{algorithmic}
\end{algorithm}

The Stable Diffusion-based Momentum Integrated Adversarial Examples (SD-MIAE) is summarized in Algorithm~\ref{algo:SD-MIAE}. It refines adversarial image generation by combining Stable Diffusion's text-to-image capabilities with a momentum-based optimization strategy. Initially, token embeddings are generated from a text prompt, and an image is synthesized using these embeddings. The algorithm then iteratively optimizes the token embeddings to craft adversarial examples that mislead a target classifier. In each iteration, the adversarial loss is computed to drive misclassification, while cosine similarity regularization ensures minimal perturbations, preserving image fidelity. Gradients of the loss are accumulated using a momentum factor, stabilizing the perturbation direction and enhancing attack robustness. This process continues until a final adversarial image is produced, effectively deceiving the classifier while maintaining the natural appearance of the image.

\section{Experiments and Results}

In this section, we conduct experiments to evaluate the effectiveness of the adversarial examples generated by the SD-MIAE. We will show the proposed SD-MIAE is able to generate adversarial examples that are highly effective in misleading the state-of-the-art image classifier while maintaining the imperceptibility of adversarial perturbations and semantic similarity to the original class label.

\subsection{Experimental Setup}

\textbf{Models.} Following~\cite{b12}, we employ a nano version of the Stable Diffusion model finetuned from the official 2.1 release (model tag: ``bguisard/stable-diffusion-nano-2-1") to generate adversarial examples.
The images generated by Stable Diffusion are initially 128$\times$128 in resolution and are resized to 224$\times$224 before being input into the target classifier, which we use ResNet-50~\cite{b24} pretrained on ImageNet in the experiments to match its default resolution. The Stable Diffusion model adopts a DDIM sampler with 20 sampling steps, and the guidance scale is set to 8.5.

\textbf{Dataset and Evaluation metrics.} We use the ImageNet-100~\cite{b23}, a widely used subset of the larger ImageNet dataset, for evaluation. It comprises 100 classes and offers a diverse set of high-resolution images. The ImageNet-100 dataset is particularly suited for evaluating adversarial attacks due to its diversity and complexity, which present significant challenges for image classifiers. To ensure a fair and meaningful evaluation, we focus on classes whose accuracy is higher than 90\% with no attack within this dataset. Specifically, we select classes where our target classifier, ResNet-50\cite{b24}, achieves at least 90\% accuracy in the absence of any adversarial perturbations. This selection process yields 25 classes, from which we choose the first 10 for our experiments: castle, flamingo, forklift, fountain, hamster, koala, knot, monarch, tiger, and zebra. The target classifier achieves 96\% of accuracy on these 10 selected classes.

The misclassification rate is used to evaluate the effectiveness of our adversarial examples. It measures the percentage of generated adversarial images that are successfully misclassified by the target classifier while still maintaining their semantic similarities to their original class labels. Specifically, we first count all misclassified samples and then manually verify if they still resemble their original class.

\textbf{Baseline setup.}  To establish a baseline for our experiments, we first generate a set of benign images for each of the ten selected classes aforementioned. For each class, 20 different random latent vectors are prepared and used to generate images with Stable Diffusion. Note that these images are generated without any adversarial optimization, and each can be correctly classified by the target classifier. This step is crucial for ensuring that the initialized samples are not already natural adversarial examples (NAEs), allowing us to isolate the effect of the optimization and attack so we can fairly attribute the adversarial properties of the images to our methodology rather than to inherent factors in the generative model. We refer to this initial step as ``benign" in our experiments, indicating that no adversarial perturbation is applied.

After generating the benign images, we proceed with 100 optimizations—10 classes and 10 prepared latent vectors per class-each corresponding to one class and one prepared latent vector. We ensure that each image is correctly classified before adding any adversarial perturbations.

Following the benign setup, we initiated the optimization process to generate adversarial examples. For each image in the dataset, a text prompt describing the target class (e.g., ``A high-quality image of a hamster") is used to initialize the token embeddings. The token embeddings associated with the class label are then iteratively optimized. The goal of this process is to subtly modify the input images generated by SD such that they would be misclassified by the target model while maintaining high visual fidelity. 
The optimization is conducted over 25 steps using the Adam optimizer~\cite{b30} with a learning rate of 0.001. 
During each iteration, we update the token embeddings and generate an image using the Stable Diffusion, similar to~\cite{b12}.

\textbf{Momentum-Based Refinement.} To further enhance the adversarial examples, we incorporate a momentum-based refinement technique introduced in Sec~\ref{method:momentum}.
The refinement is performed by a joint optimization of epsilon ($\epsilon$) and momentum ($\mu$). It is conducted over 30 iterations, testing $\epsilon$ in [0.0, 2.0] with intervals of 0.1 and $\mu$ in [0.5, 1.5] with intervals of 0.2 to identify the optimal settings. Through systematic trials, we set $\epsilon=0.2$ and $\mu=1.0$ to achieve the highest misclassification rate while preserving the images' visual fidelity.


\subsection{Results and Analysis}

\begin{figure}[t]
    \centering
    \includegraphics[width=0.85\linewidth]{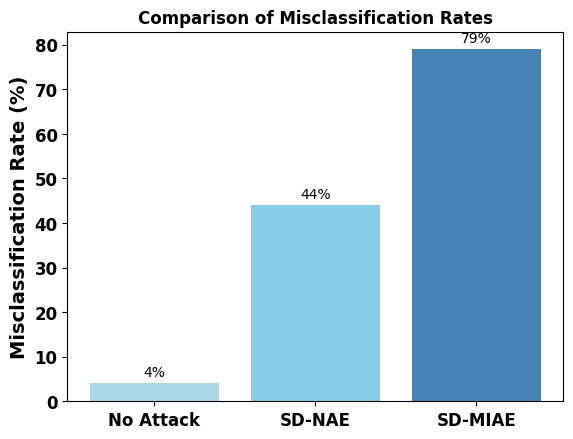}
    \caption{Comparison of the misclassification rates across 100 images from 10 classes on the benign setup (4\%), SD-NAE setup (44\%), SD-MIAE setup (79\%).}
    \label{fig:misclassification_comparison}
\end{figure}

\begin{figure*}[t]
    \centering
    \includegraphics[width=0.9\textwidth]{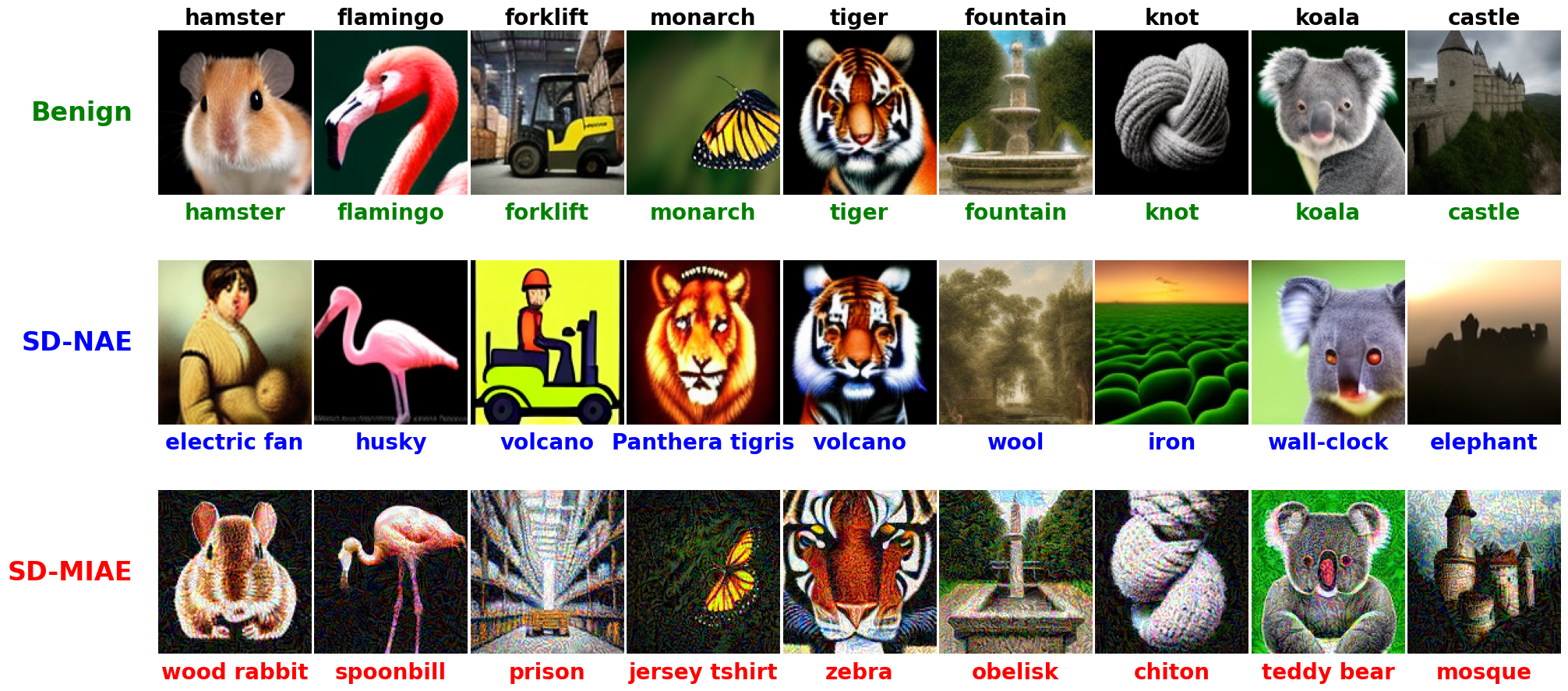}
    \caption{Visualization of the impact of various attack modes on the classification of images. Each row corresponds to a different attack mode: (1) benign (unaltered images), (2) SD-NAE, and (3) SD-MIAE. Columns represent different classes, with predicted class labels shown below each image. Note that adversarial examples generated by SD-MIAE maintain semantic similarity to the original class, while those from SD-NAE deviate from it.
    }
    \label{fig:attack_modes}
\end{figure*}

\textbf{Main results.} \textbf{Figure \ref{fig:misclassification_comparison}} presents the misclassification rate compared to the baseline method. The proposed SD-MIAE achieves 79\% misclassification rate, improving by 35\% over SD-NAE. This suggests the improved effectiveness of SD-MIAE in generating adversarial examples that can successfully mislead the target classifier. More importantly, SD-MIAE can better preserve the \textbf{semantic similarity of generated adversarial examples to their original class label}. \textbf{Figure \ref{fig:attack_modes}} visualizes benign images generated by Stable Diffusion alongside adversarial examples from SD-NAE and our proposed SD-MIAE. The results demonstrate that SD-MIAE can generate adversarial examples that not only effectively mislead the classifier but also preserve the semantic similarity to the original class label. In contrast, SD-NAE produces examples that deviate significantly from the original labels. For instance, SD-NAE generates a human, a tiger, and a forest for the classes ``hamster", ``monarch", and ``fountain", respectively. These semantic deviations are easily detectable and filterable, while adversarial samples generated by SD-MIAE maintain correct class associations.

\begin{figure}[t]
    \centering
\includegraphics[width=0.90\linewidth]{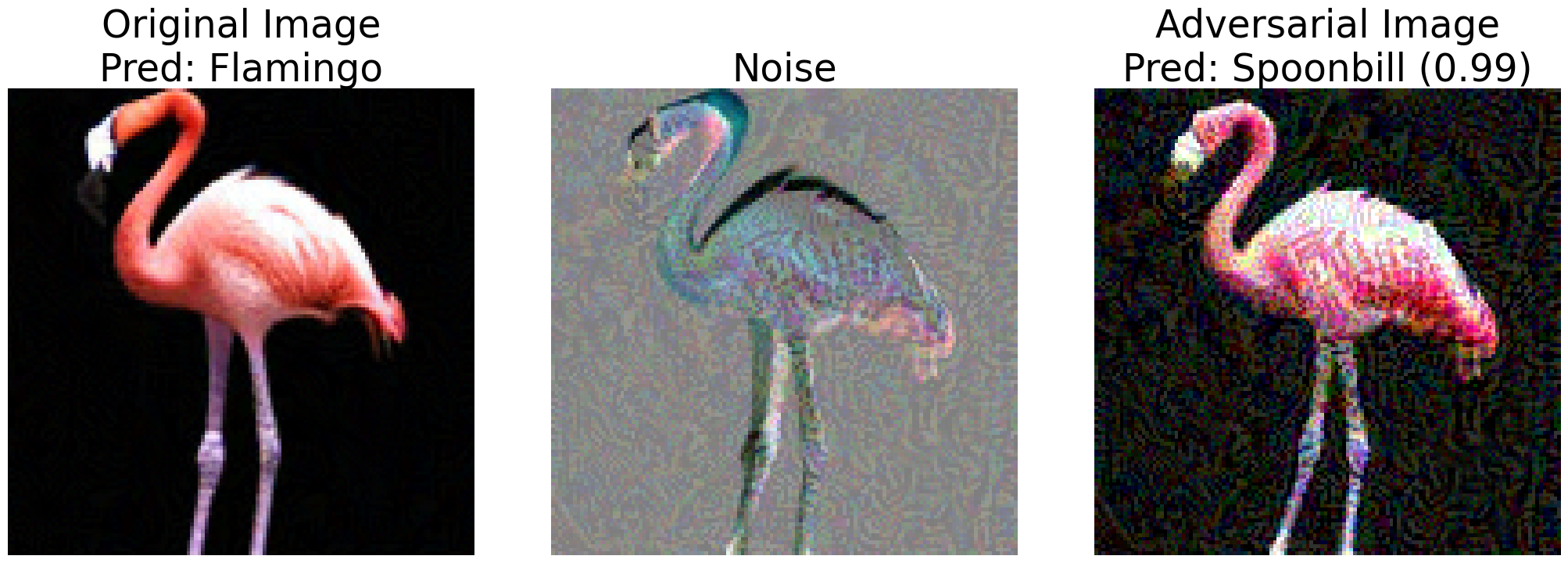}
    \caption{Visualization of the effects of the SD-MIAE attack on a benign image. The left image shows the original image classified as a Flamingo. The middle image depicts the adversarial perturbations optimized by our momentum-based refinement technique. The right image displays the adversarial image misclassified as a Spoonbill with a probability of 99\%.}
    \label{fig:mifgsm_effects}
\end{figure}

Note that while a higher $\epsilon$ increases the misclassification rate, we set $\epsilon=0.2$ conservatively since a higher $\epsilon$ also introduces more visible artifacts in the images. This trade-off between adversarial effectiveness and image quality is carefully managed to maintain both a high misclassification rate and imperceptibility.  \textbf{Figure~\ref{fig:mifgsm_effects}} shows the effects of the SD-MIAE attack on a benign image (Flamingo) with $\epsilon=0.2$ and $\mu = 1.0$. It can be seen that the adversarial image maintains its natural appearance as the benign image after applying the adversarial perturbations. In addition, the target classifier now predicts this sample as Spoonbill with a high confidence of 0.99, demonstrating SD-MIAE's ability to generate highly effective adversarial examples. 

\begin{figure}[b]
    \centering
    \includegraphics[width=0.85\linewidth]{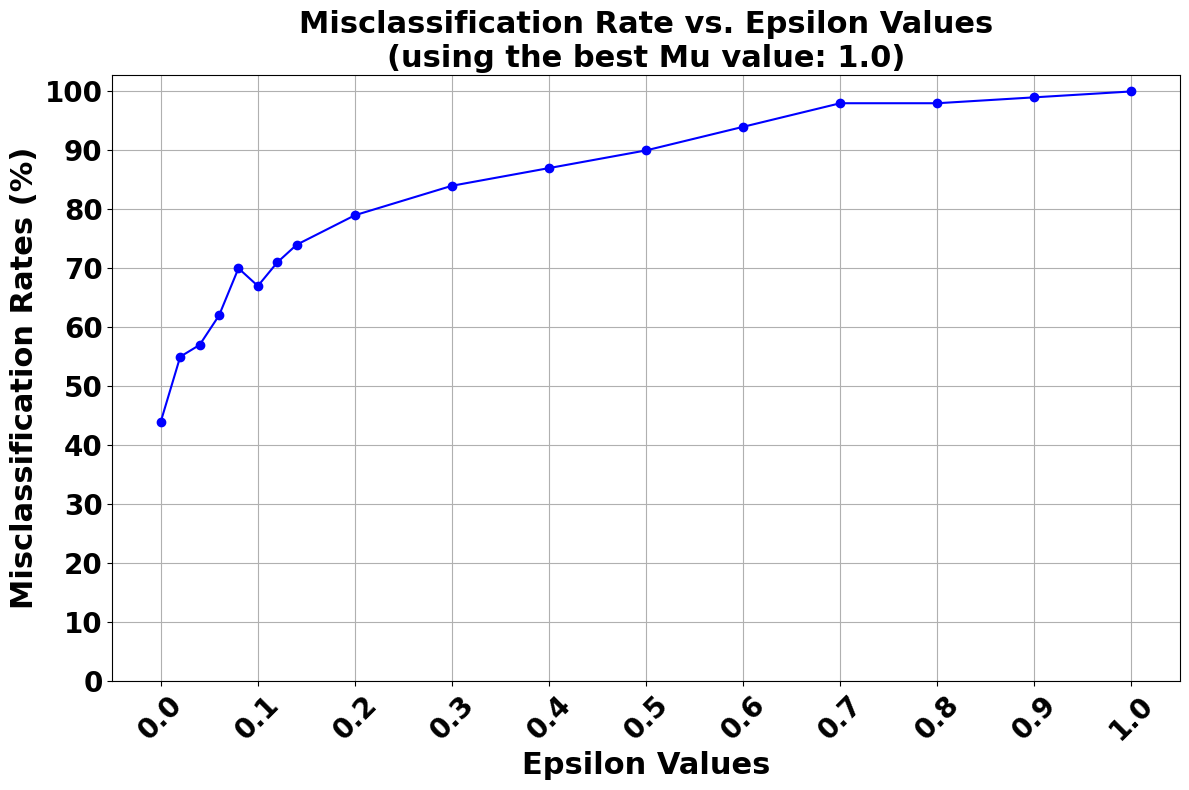}
    \caption{Impact of increasing epsilon using the best $\mu$, 1.0, on the misclassification rate.}
    \label{fig:epsilon_mu_1.0_impact_missrate}
\end{figure}

\begin{figure}[t]
    \centering
    \includegraphics[width=0.9\linewidth]{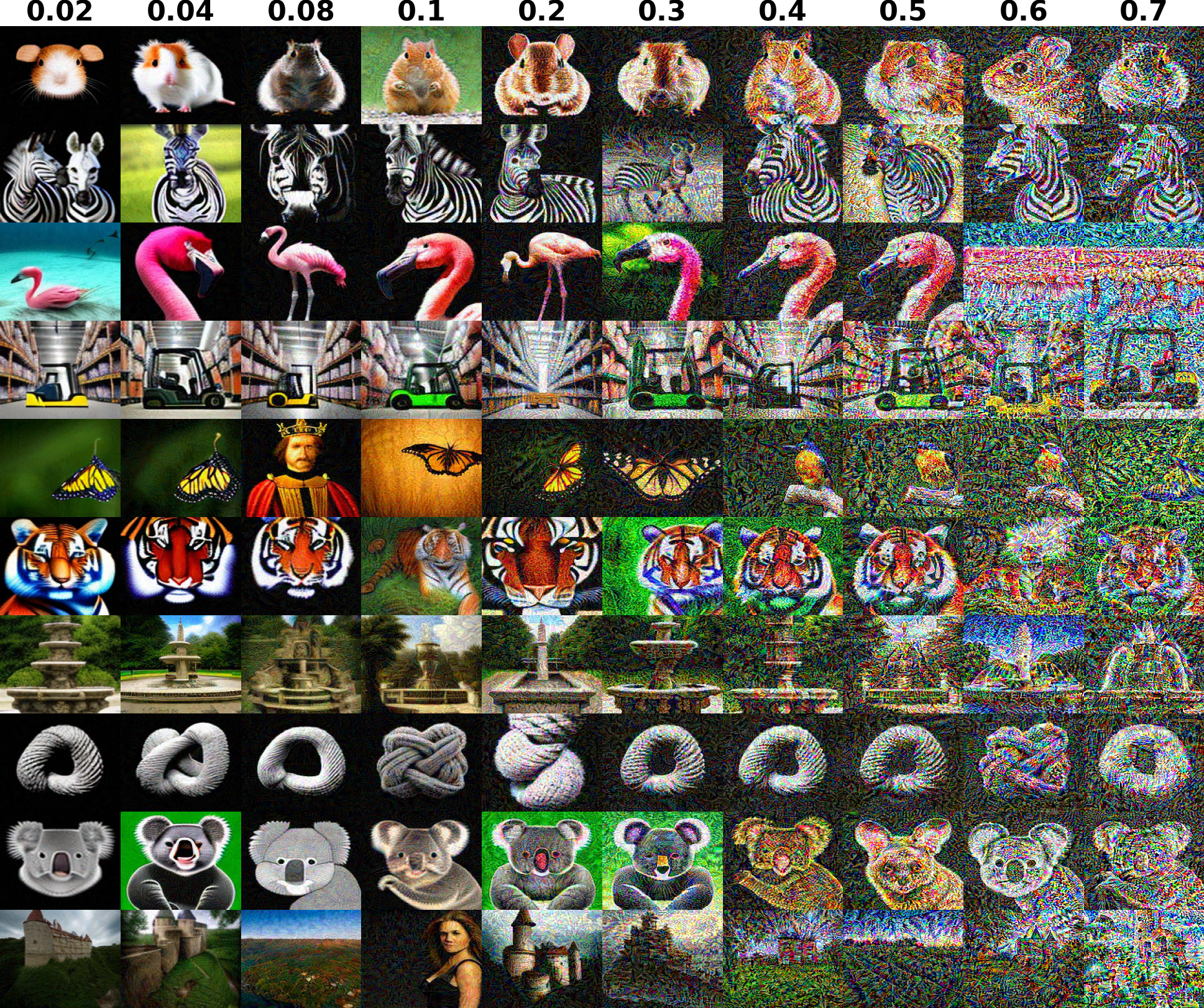}
    \caption{Visulization of generated adversarial examples with increasing $\epsilon$ using the best $\mu$, 1.0.}
    \label{fig:epsilon_mu_1.0_impact_visualization}
\end{figure}

Additionally, the combination of $\epsilon$ and $\mu$ in the SD-MIAE framework is critical. \textbf{Figure~\ref{fig:epsilon_mu_1.0_impact_missrate}} illustrates the misclassification rate with varying epsilon at a fixed $\mu=1.0$, while \textbf{Figure~\ref{fig:epsilon_mu_1.0_impact_visualization}} visualizes the resulting adversarial examples. As seen, $\epsilon$ controls the perturbation magnitude, directly affecting the attack's success, while $\mu$ stabilizes and guides the optimization of these perturbations across iterations, ensuring consistency and preserving the image's natural appearance. This balance enables the generation of adversarial examples that are both highly effective in misleading the classifier and visually indistinguishable from their original counterparts.

\begin{figure}[t]
    \centering
    \includegraphics[width=0.85\columnwidth]{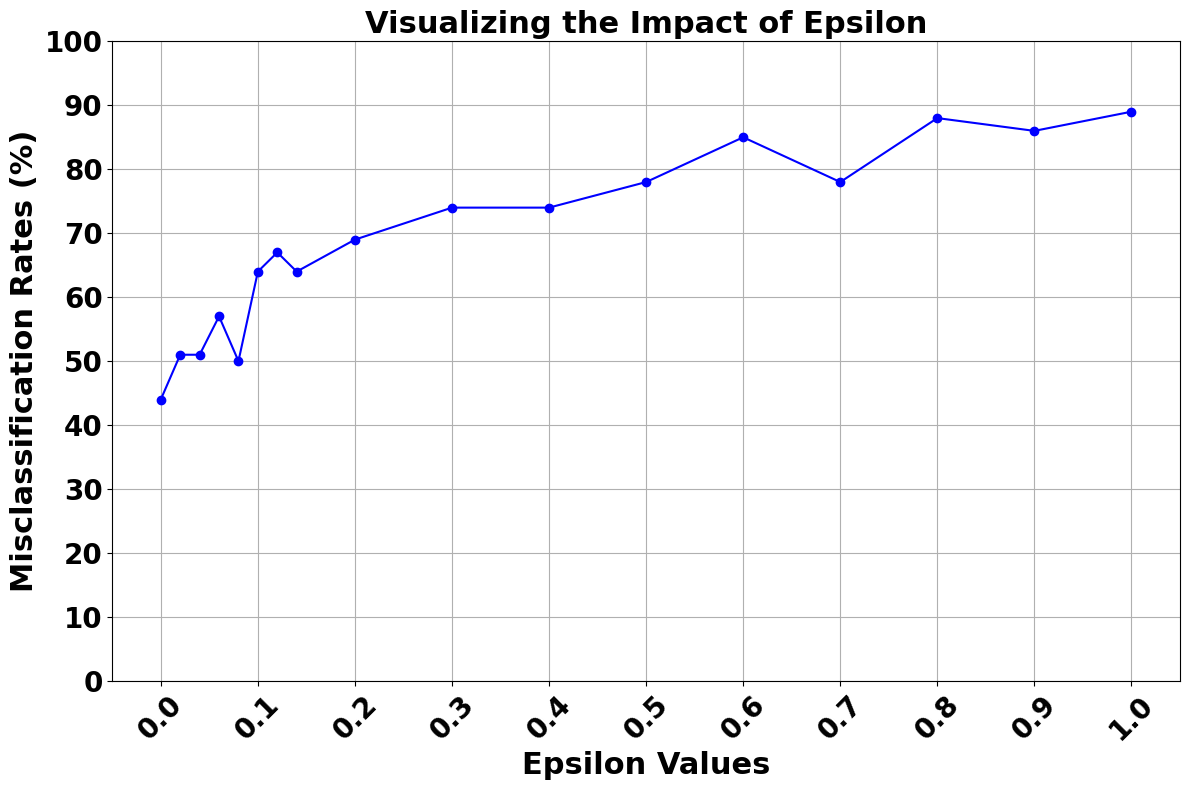}
    \caption{Impact of epsilon on the misclassification rate.}
    \label{fig:epsilon_impact}
\end{figure}
\begin{figure}[t]
    \centering
    \includegraphics[width=0.9\linewidth]{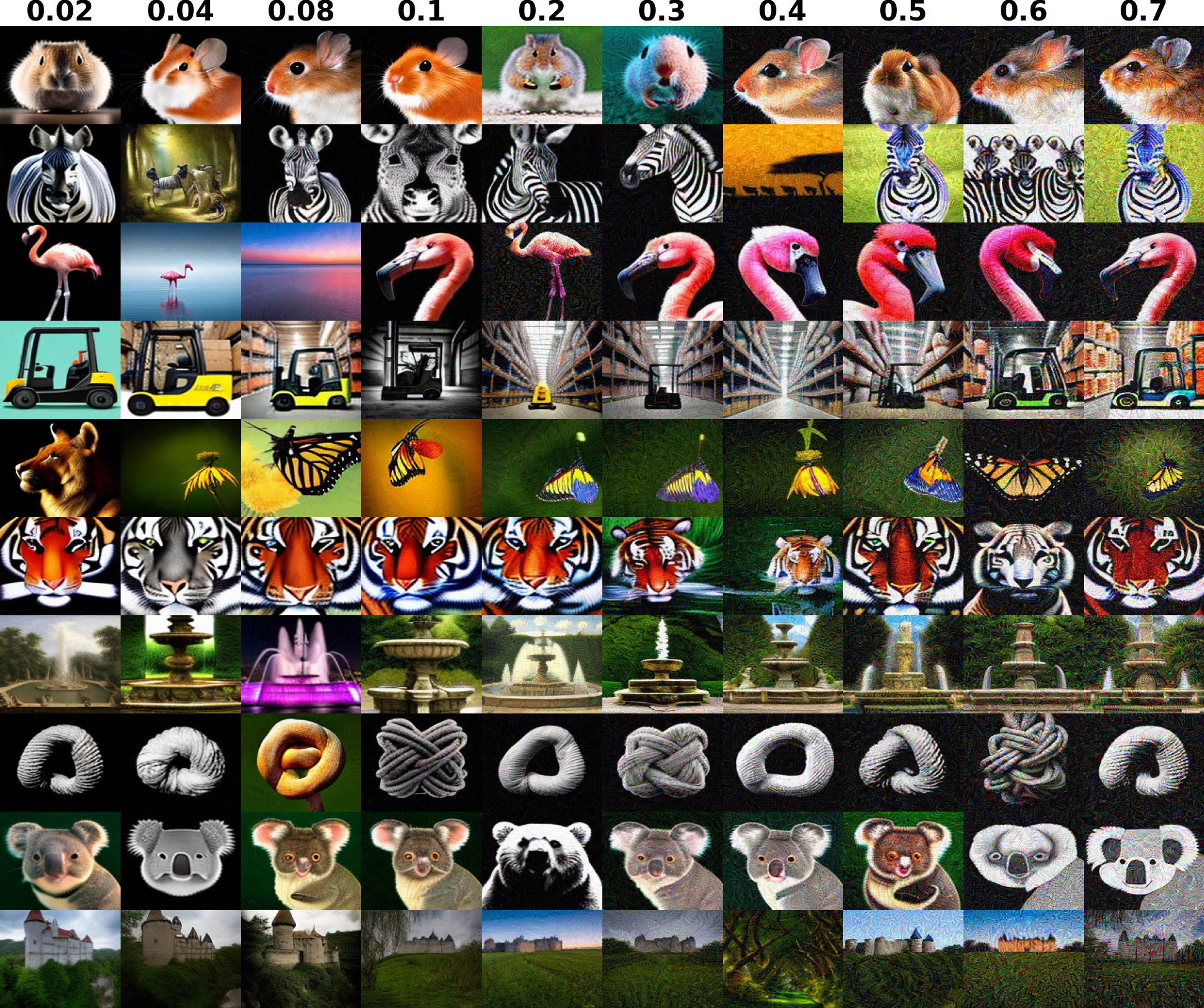}
    \caption{Visualization of adversarial examples generated by varying $\epsilon$. $\mu$ is set to 0. The increase of $\epsilon$ induces more perceptible pertubations.}
    \label{fig:epsilon_visualization}
\end{figure}

\textbf{Impact of Epsilon.} To investigate the impact and limitations of relying solely on $\epsilon$ for adversarial attacks, \textbf{Figure~\ref{fig:epsilon_impact}} presents the misclassification rate as $\epsilon$ varies with momentum factor $\mu=0$. As $\epsilon$ increases from 0.0 to 1.0, the misclassification rates rise significantly, from 44\% to 89\%. Notably, even at $\epsilon=1.0$, it can only achieve 89\% of misclassification rate,  while in \textbf{Figure~\ref{fig:epsilon_mu_1.0_impact_missrate}} it can achieve 100\%. This highlights the effectiveness of adding the momentum term when optimizing the adversarial perturbation. However, increasing $\epsilon$ also makes the perturbations more perceptible and thus reduces the visual quality, especially at higher levels. As illustrated in \textbf{Figure~\ref{fig:epsilon_visualization}}, while the images generated with higher $\epsilon$ do not become completely filled with noise, they exhibit noticeable artifacts that degrade their visual fidelity. This underscores the importance of balancing $\epsilon$ to maintain image quality while achieving effective attacks.

\begin{figure}[t]
    \centering
    \includegraphics[width=0.85\linewidth]{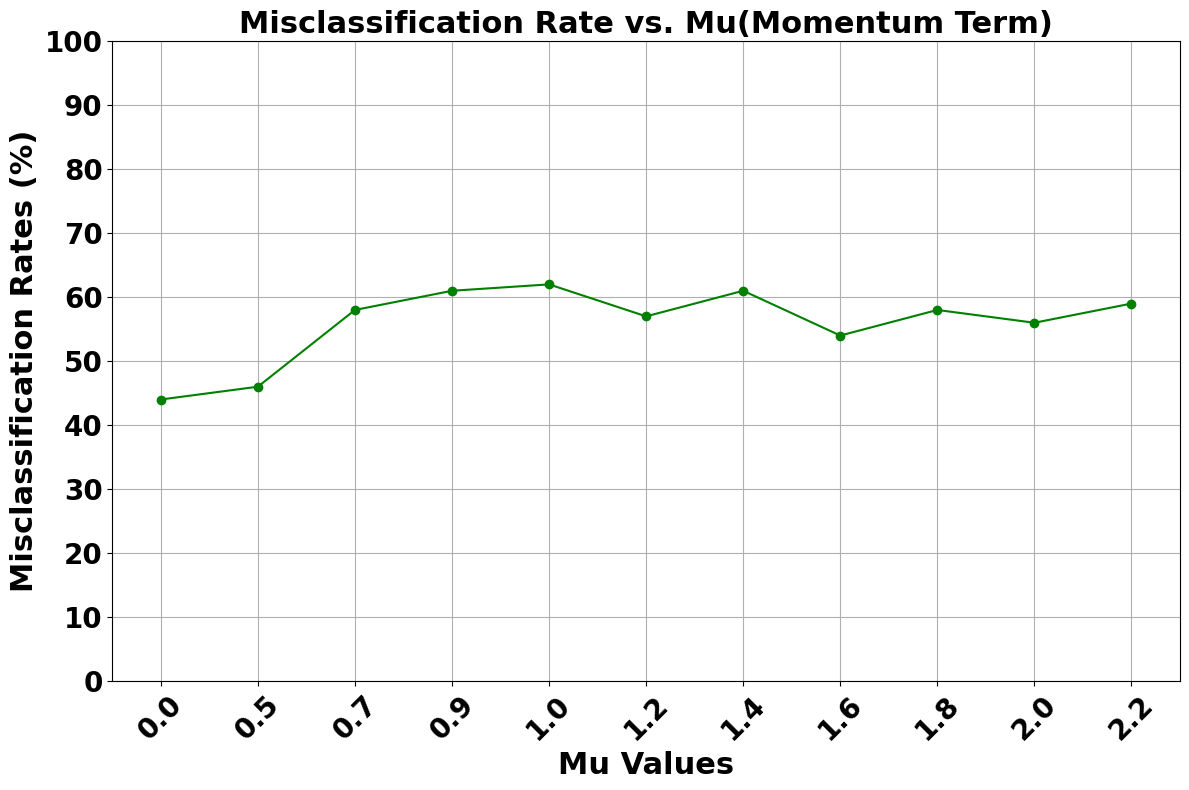}
    \caption{Impact of momentum ($\mu$) on the misclassification rate.}
    \label{fig:mu_impact}
\end{figure}

\textbf{Impact of momentum.} We further analyze the impact of the momentum factor on the effectiveness of the generated adversarial examples by fixing $\epsilon$ to 0.03 and gradually increasing the value of $\mu$. \textbf{Figure~\ref{fig:mu_impact}} shows the misclassification rate as $\mu$ varies from 0.0 to 2.2. It can be seen that as $\mu$ increases from 0.0 to 1.0, the misclassification improves from 44\% to 62\%, demonstrating that increasing momentum factor can significantly enhance the effectiveness of the adversarial example, even when $\epsilon$ is set to a small value as 0.03. 


Note that the misclassification rate plateaus at $\mu=1.0$, with no further improvement beyond this point. As shown in Equation~\ref{equa:momentum}$,~\mu=1.0$ means that the current update is performed by adding up all previous gradients. Further scaling up the sum of all gradients no longer contributes to the effectiveness of adversarial examples generation.

\begin{figure}[t]
    \centering
    \includegraphics[width=1.0\linewidth]{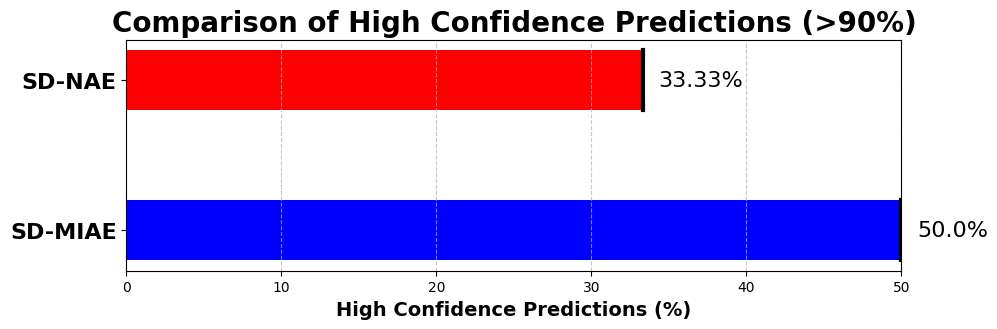}
    \caption{SD-MIAE shows greater effectiveness in generating confidently incorrect predictions, with 50\% of misclassifications having over 90\% prediction confidence, compared to 33.33\% for SD-NAE.}
    \label{fig:confidence_comparison}
\end{figure}

Finally, as shown in \textbf{Figure \ref{fig:confidence_comparison}}, we compare the confidence of incorrect predictions caused by adversarial examples generated by SD-NAE and SD-MIAE. Particularly, SD-MIAE shows greater effectiveness in generating
confidently incorrect predictions, with 50\% of misclassifications having over 90\% prediction confidence, compared to 33.33\% for SD-NAE. Notably, for SD-NAE, such high confidence occurs mainly when adversarial images generated by it are highly distorted and unrecognizable. Additionally, the results also suggest that the adversarial examples generated by SD-MIAE exploit more fundamental flaws in the decision boundaries of neural networks\cite{b17}, indicating that SD-MIAE-generated adversarial examples are more challenging for both neural network classifiers and human observers to identify. 

\section{Discussion}

\begin{figure}[t]
    \centering
    \includegraphics[width=1.0\linewidth]{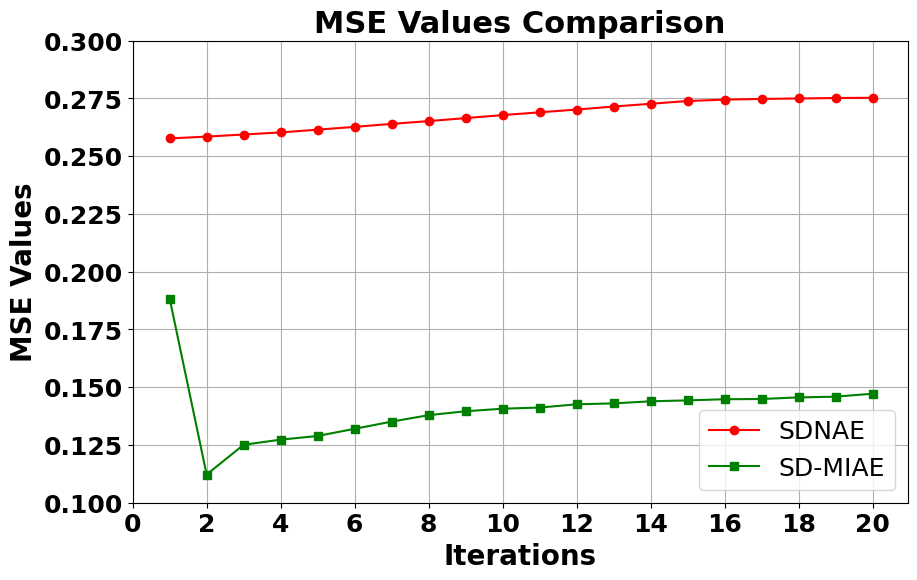}
    \caption{Mean Squared Error (MSE) Between Original Images and Adversarial Examples Generated by SD-NAE and SD-MIAE Across Iterations.}
    \label{fig:mse_values}
\end{figure}

While the Stable Diffusion-based Momentum Integrated Adversarial Examples (SD-MIAE) framework presents significant advancements in generating more effective and natural adversarial examples, several limitations must be acknowledged: (1) Targeted attacks remain challenging. We also conduct targeted attack experiments. Specifically, we set the adversarial target label to their $\text{\textit{class\_id}} - 1 $ (e.g., if the \textit{class\_id} for 'hamster' is 333, the adversarial target is set to 332) and $\epsilon=0.2$ and $\mu=1$ for SD-MIAE. We conduct these experiments for both SD-NAE and SD-MIAE, achieving misclassification rates of 0\% and 7\%, respectively. Recall that for a successful targeted attack, the image must be misclassified into the specific adversarial target label while maintaining semantic similarity to the original class label. The requirement to force classification into a specific class often results in excessive perturbation of the class token embedding and thus causes generated images to deviate significantly from their original class labels, leading to low misclassification rates. However, SD-MIAE still outperforms SD-NAE by 7\%. (2) The framework's reliance on specific configurations, namely the Stable Diffusion model and the ResNet-50 classifier. These models are well-regarded in the field, but their use raises questions about the generalizability of SD-MIAE to other architectures and application domains. Future work should explore the adaptability of SD-MIAE across different generative models and classifiers to evaluate its broader applicability and robustness.
(3) Tuning epsilon, which controls perturbation magnitude, and the momentum term is essential for achieving effectiveness. Our results show that these parameters can be adjusted to maximize misclassification rates while preserving visual fidelity. However, finding the optimal balance between perturbation strength and image quality is non-trivial. Future research should further explore the dynamic interaction between these parameters. Developing adaptive algorithms to automatically optimize these settings across different scenarios would significantly enhance the robustness and versatility of SD-MIAE.
(4) The trade-off between computational overhead and effective adversarial sample generation of SD-MIAE also presents a limitation, particularly in terms of GPU memory and processing time. Generating a single 128x128 adversarial example requires $ \sim $22 GB of GPU memory and takes around 37 seconds per sample, depending on the hardware configuration and number of iterations. In comparison, SD-NAE takes 17 seconds, while the original Stable Diffusion takes only 2 seconds. Compared to SD-NAE, the additional time for SD-MIAE is due to the momentum-based refinement process, which is crucial for producing more effective adversarial samples as validated in our experiments. However, this computational overhead may limit its scalability without access to high-performance computational resources.
(5) The imperceptibility of adversarial perturbations is crucial for evaluating the effectiveness of adversarial examples. Our experiments indicate that SD-MIAE generates perturbations that are less perceptible compared to those produced by SD-NAE. As shown in \textbf{Figure \ref{fig:mse_values}}, the Mean Squared Error (MSE) between the original images and the adversarial examples is consistently lower for SD-MIAE across various iterations. The lower MSE values for SD-MIAE suggest that the perturbations it introduces are subtler, thereby making the adversarial examples more visually similar to the original images. This enhanced imperceptibility is essential for adversarial attacks to remain undetected. Future research should delve deeper into evaluating imperceptibility using a variety of perceptual metrics to further validate and improve the stealthiness of adversarial examples generated by SD-MIAE.

Furthermore, while SD-MIAE demonstrates improvements in maintaining visual semantic coherence and misclassification rate, further research is needed to evaluate its performance against advanced detection systems. As adversarial detection evolves, future studies should assess SD-MIAE's resilience against various detection strategies and explore its applicability across different data modalities, such as 3D data, to fully assess its potential\cite{b36}.

\section{Conclusion}

In this work, we introduced Stable Diffusion-based Momentum Integrated Adversarial Examples (SD-MIAE), a novel framework that significantly improves the generation of adversarial examples by incorporating momentum-based optimization into the Stable Diffusion model. SD-MIAE achieves a misclassification rate of 79\%, improving by 35\% over the
state-of-the-art method, demonstrating its ability to generate highly effective adversarial examples that mislead the classifier. Additionally, adversarial examples generated by SD-MIAE also preserve the visual imperceptibility and the semantic similarity to the original class label, making it a practical method for robust
adversarial evaluation. 




\bibliographystyle{IEEEtran}

\begin{thebibliography}{10}
\providecommand{\url}[1]{#1}
\csname url@samestyle\endcsname
\providecommand{\newblock}{\relax}
\providecommand{\bibinfo}[2]{#2}
\providecommand{\BIBentrySTDinterwordspacing}{\spaceskip=0pt\relax}
\providecommand{\BIBentryALTinterwordstretchfactor}{4}
\providecommand{\BIBentryALTinterwordspacing}{\spaceskip=\fontdimen2\font plus
\BIBentryALTinterwordstretchfactor\fontdimen3\font minus \fontdimen4\font\relax}
\providecommand{\BIBforeignlanguage}[2]{{%
\expandafter\ifx\csname l@#1\endcsname\relax
\typeout{** WARNING: IEEEtran.bst: No hyphenation pattern has been}%
\typeout{** loaded for the language `#1'. Using the pattern for}%
\typeout{** the default language instead.}%
\else
\language=\csname l@#1\endcsname
\fi
#2}}
\providecommand{\BIBdecl}{\relax}
\BIBdecl

\bibitem{b1}
A.~Krizhevsky, I.~Sutskever, and G.~E. Hinton, ``Imagenet classification with deep convolutional neural networks,'' in \emph{Advances in Neural Information Processing Systems}, vol.~25, 2012, pp. 1097--1105.

\bibitem{b2}
G.~Hinton \emph{et~al.}, ``Deep neural networks for acoustic modeling in speech recognition,'' \emph{IEEE Signal Processing Magazine}, vol.~29, no.~6, pp. 82--97, 2012.

\bibitem{b3}
J.~Devlin, M.~W. Chang, K.~Lee, and K.~Toutanova, ``Bert: Pre-training of deep bidirectional transformers for language understanding,'' in \emph{Proc. 2019 Conf. North American Chapter of the Association for Computational Linguistics: Human Language Technologies}, 2019, pp. 4171--4186.

\bibitem{b4}
I.~J. Goodfellow, J.~Shlens, and C.~Szegedy, ``Explaining and harnessing adversarial examples,'' in \emph{Proc. Int. Conf. Learning Representations (ICLR)}, 2015.

\bibitem{b5}
S.~Karanam, ``Safety and robustness of autonomous vehicles,'' \emph{Journal of Machine Learning Research}, vol.~21, pp. 1--28, 2020.

\bibitem{b6}
A.~Esteva \emph{et~al.}, ``Dermatologist-level classification of skin cancer with deep neural networks,'' \emph{Nature}, vol. 542, no. 7639, pp. 115--118, 2017.

\bibitem{b19}
R.~Rombach, A.~Blattmann, D.~Lorenz, P.~Esser, and B.~Ommer, ``High-resolution image synthesis with latent diffusion models,'' 2021.

\bibitem{b27}
C.~Zhang, M.~Hua, W.~Li, and L.~Wang, ``Adversarial attacks and defenses on text-to-image diffusion models: A survey,'' \emph{arXiv preprint arXiv:2407.15861}, 2024.

\bibitem{b28}
A.~Hila, T.~Nguyen, and J.~Zhu, ``Robustness of text-to-image diffusion models against adversarial attacks,'' \emph{Proceedings of the IEEE/CVF Conference on Computer Vision and Pattern Recognition (CVPR)}, 2023.

\bibitem{b12}
Y.~Lin, J.~Zhang, Y.~Chen, and H.~Li, ``Sd-nae: Generating natural adversarial examples with stable diffusion,'' 2024.

\bibitem{b16}
N.~Carlini and D.~Wagner, ``Towards evaluating the robustness of neural networks,'' in \emph{Proc. IEEE Symp. Security and Privacy (SP)}, 2017.

\bibitem{b33}
K.~Eykholt, I.~Evtimov, E.~Fernandes, B.~Li, A.~Rahmati, C.~Xiao, A.~Prakash, T.~Kohno, and D.~Song, ``Robust physical-world attacks on deep learning visual classification,'' in \emph{IEEE Conference on Computer Vision and Pattern Recognition (CVPR)}, 2018.

\bibitem{b14}
A.~Madry, A.~Makelov, L.~Schmidt, D.~Tsipras, and A.~Vladu, ``Towards deep learning models resistant to adversarial attacks,'' in \emph{Proc. Int. Conf. Learning Representations (ICLR)}, 2018.

\bibitem{b8}
I.~J. Goodfellow, J.~Shlens, and C.~Szegedy, ``Explaining and harnessing adversarial examples,'' \emph{arXiv:1412.6572}, 2015.

\bibitem{b20}
T.~B. Brown, D.~Mane, A.~Roy, M.~Abadi, and J.~Gilmer, ``Adversarial patch,'' \emph{arXiv:1712.09665}, 2017.

\bibitem{b21}
W.~Xu, D.~Evans, and Y.~Qi, ``Feature squeezing: Detecting adversarial examples in deep neural networks,'' 2017.

\bibitem{b29}
\BIBentryALTinterwordspacing
S.~Goyal, S.~Doddapaneni, M.~M. Khapra, and B.~Ravindran, ``A survey of adversarial defenses and robustness in nlp,'' \emph{ACM Computing Surveys}, vol.~55, no. 14s, July 2023. [Online]. Available: \url{https://doi.org/10.1145/3543873}
\BIBentrySTDinterwordspacing

\bibitem{b9}
D.~Hendrycks, K.~Song, and S.~Basart, ``Natural adversarial examples,'' \emph{OpenReview}, 2021.

\bibitem{b15}
Y.~Song, T.~Kim, S.~Nowozin, S.~Ermon, and N.~Kushman, ``Pixeldefend: Leveraging generative models to understand and defend against adversarial examples,'' in \emph{Proc. Int. Conf. Learning Representations (ICLR)}, 2018.

\bibitem{b34}
A.~Vaswani, N.~Shazeer, N.~Parmar, J.~Uszkoreit, L.~Jones, A.~N. Gomez, {\L}.~Kaiser, and I.~Polosukhin, ``Attention is all you need,'' in \emph{Advances in Neural Information Processing Systems (NeurIPS)}, vol.~30, 2017, pp. 5998--6008.

\bibitem{dhariwal2021diffusion}
P.~Dhariwal and A.~Nichol, ``Diffusion models beat gans on image synthesis,'' in \emph{Advances in Neural Information Processing Systems}, vol.~34, 2021, pp. 8780--8794.

\bibitem{ho2021classifier}
J.~Ho and T.~Salimans, ``Classifier-free diffusion guidance,'' in \emph{NeurIPS 2021 Workshop on Deep Generative Models and Downstream Applications}, 2021.

\bibitem{b7}
Y.~Dong, S.~Du, and H.~Xu, ``Improving the robustness of deep neural networks via momentum iterative training,'' \emph{arXiv:1710.06081}, 2018.

\bibitem{b24}
K.~He, X.~Zhang, S.~Ren, and J.~Sun, ``Deep residual learning for image recognition,'' in \emph{Proc. IEEE Conf. Computer Vision and Pattern Recognition (CVPR)}, 2016, pp. 770--778.

\bibitem{b23}
O.~Russakovsky \emph{et~al.}, ``Imagenet large scale visual recognition challenge,'' \emph{Int. J. Comput. Vision}, vol. 115, no.~3, pp. 211--252, 2015.

\bibitem{b30}
D.~P. Kingma and J.~Ba, ``Adam: A method for stochastic optimization,'' \emph{arXiv preprint arXiv:1412.6980}, 2014.

\bibitem{b17}
C.~Szegedy \emph{et~al.}, ``Intriguing properties of neural networks,'' \emph{arXiv:1312.6199}, 2013.

\bibitem{b36}
W.~Tu, Y.~Wang, X.~Liu, Y.~Chen, Y.~Gao, Y.~Fan, and H.~Xue, ``Physically realizable adversarial examples for lidar object detection,'' in \emph{Proceedings of the IEEE/CVF Conference on Computer Vision and Pattern Recognition (CVPR)}, 2020, pp. 1361--1370.

\end{thebibliography}

\end{document}